\documentclass[runningheads,a4paper]{llncs}

\usepackage{amssymb}
\usepackage{graphicx}
\usepackage{indentfirst}

\usepackage{multirow}
\usepackage{array}
\newcolumntype{C}[1]{>{\centering\let\newline\\\arraybackslash\hspace{0pt}}m{#1}}
\usepackage{verbatim}

\usepackage{url}
\urldef{\mailsa}\path|iwyoo@unist.ac.kr|
\urldef{\mailsb}\path|anna.kramer, leonie.kunz, christine.reiss, nicole.sator,|
\urldef{\mailsc}\path|erika.siebert-cole, peter.strasser, lncs}@springer.com| 

\usepackage[usenames,dvipsnames]{color}
\definecolor{cb_orange}{rgb}{1.0,0.51,0.0}
\definecolor{cb_blue}{rgb}{0.22,0.49,0.72}
\definecolor{cb_green}{rgb}{0.3,0.67,0.29}
\definecolor{cb_red}{rgb}{0.89,0.1,0.11}
\definecolor{cb_purple}{rgb}{0.6, 0.31, 0.64}

\newcommand{\todo}[1]{{\textcolor{cb_red}{TODO}}}

\begin{document}

\mainmatter  

\title{ssEMnet: Serial-section Electron Microscopy Image Registration using a Spatial Transformer Network with Learned Features}

\titlerunning{ssEM Image Registration using a STN with Learned Features}


\author{
Inwan Yoo$^1$, David G. C. Hildebrand$^2$, Willie F. Tobin$^3$, \\ 
Wei-Chung Allen Lee$^3$ and Won-Ki Jeong$^1$
}


\institute{
Ulsan National Institute of Science and Technology$^1$\\
The Rockefeller University$^2$\\
Harvard Medical School$^3$ \\
E-mail: \mailsa\\
}


\toctitle{Lecture Notes in Computer Science}
\tocauthor{Authors' Instructions}
\maketitle

\vspace{-0.15in}

\begin{abstract}
%
The alignment of serial-section electron microscopy (ssEM) images is critical for efforts in neuroscience that seek to reconstruct neuronal circuits. 
However, each ssEM plane contains densely packed structures that vary from one section to the next, which makes matching features across images a challenge.
Advances in deep learning has resulted in unprecedented performance in similar computer vision problems, but to our knowledge, they have not been successfully applied to ssEM image co-registration.
In this paper, we introduce a novel deep network model that combines a spatial transformer for image deformation and a convolutional autoencoder for unsupervised feature learning for robust ssEM image alignment. 
This results in improved accuracy and robustness while requiring substantially less user intervention than conventional methods.  
We evaluate our method by comparing registration quality across several datasets.
\end{abstract}


\section{Introduction}
\label{intro}

%
%
%
Ambitious efforts in neuroscience---referred to as ``connectomics''---seek to generate comprehensive brain connectivity maps.
This field utilizes the high resolution of electron microscopy (EM) to resolve neuronal structures such as dendritic spine necks and synapses, which are only tens of nanometers in size~\cite{Helmstaedter2013}.
A standard procedure for obtaining such datasets is cutting brain tissue into 30$-$50 nm-thick sections (e.g. ATUM~\cite{Hayworth2014}), acquiring images with 2$-$5 nm lateral resolution for each section, and aligning two-dimensional (2D) images into three-dimensional (3D) volumes.
Though the tissue is chemically fixed and embedded in epoxy resin to preserve ultrastructure, several deformations occur in this serial-section EM (ssEM) process.
These include tissue shrinkage, compression or expansion during sectioning, and warping from sample heating or charging due to the electron beam.
Overcoming such non-linear distortions are necessary to reproduce a 3D image volume in a state as close as possible to the original biological specimen.
Therefore, excellent image alignment is an important prerequisite for subsequent analysis.
%

%
%
Significant research efforts in image registration have been made to address medical imaging needs. 
However, ssEM image registration remains challenging due to its image characteristics: large and irregular tissue deformations with artifacts such as dusts and folds, drifting for long image sequences alignment, and difficulty in finding the optimal alignment parameters.
Several open-source ssEM image registration tools are available, such as bUnwarpJ~\cite{arganda2006consistent} and Elastic alignment~\cite{saalfeld2012elastic} (available via TrakEM2~\cite{Cardona2012a}). 
They partially address the above issues, but some of them still remain, such as lack of global regularization and complicated parameter tuning.
%

%
%
Our work is motivated by recent advances in deep neural networks.
Convolutional neural networks (CNNs) and their variants have shown unprecedented potential by largely outperforming conventional computer vision algorithms using hand-crafted feature descriptors, but their application to ssEM image registration has not been explored. 
Wu et al.~\cite{wu2015scalable} used a 3D autoencoder to extract features from MRI volumes, which are then combined with a conventional sparse, feature-driven registration method. 
Recent work by Jaderberg et al.~\cite{jaderberg2015spatial} on the spatial transformer network (STN) uses a differentiable network module inside a CNN to 
overcome the drawbacks of CNNs (i.e., lack of scale- and rotation-invariance).
Another interesting application of deep neural networks is energy optimization using backpropagation, as shown in the neural artistic style transfer proposed by Gatys et al.~\cite{gatys2016image}. 
%

%
%
Inspired by these studies, we propose a novel deep network model that is specifically designed for ssEM image registration. 
The proposed model is a novel combination of an STN and a convolutional autoencoder that generates a deformation map (i.e., vector map) for the entire image alignment via backpropagation of the network. 
We propose a feature-based image similarity measure, which is learned from the training images in an unsupervised fashion by the autoencoder. 
Unlike other conventional hand-crafted features, such as SIFT and block-matching, the learned features used in our method significantly reduce the required user parameters and make the method easy to use and less error-prone.
%
%
To the best of our knowledge, this is the first data-driven ssEM image registration method based on deep learning, which can easily extend to various applications by employing different feature encoding networks. 

\vspace{-0.07in}

\section{Method}
\label{method}

\subsection{Feature Generation using a Convolutional Autoencoder}

\begin{figure}[htb]
	\centering
	\includegraphics[width=\linewidth]{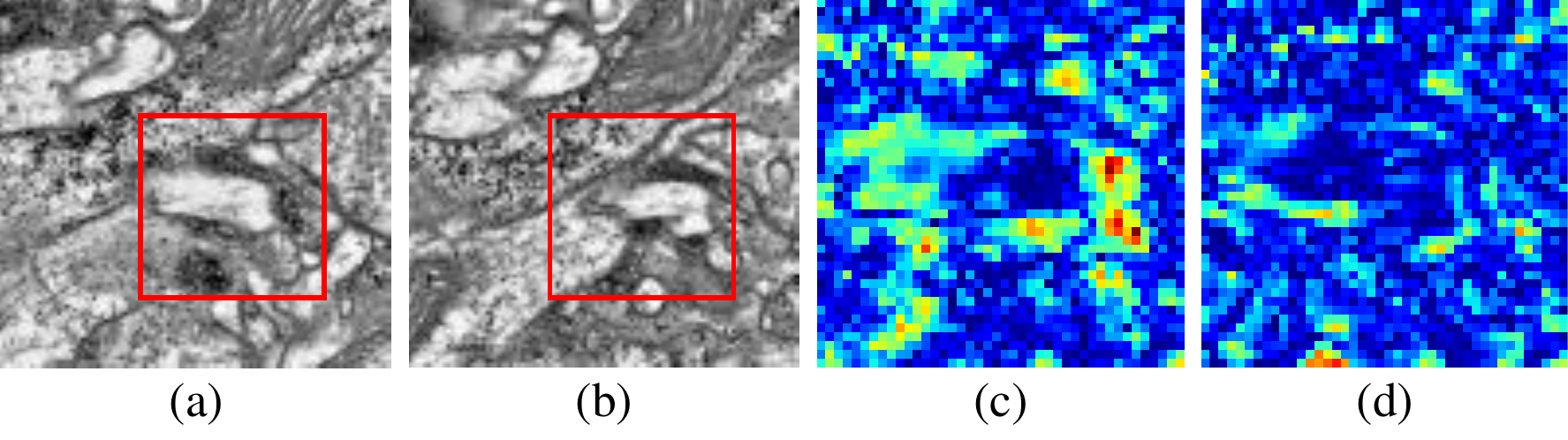}
	\caption{Comparison between the pixel intensity-based and the autoencoder feature-based registration with backpropagation. (a) the fixed image, (b) the moving image, (c) the heat map of NCC of the pixel intensity-based registration result (NCC : 0.1670), and (d) the heat map of NCC of the autoencoder feature-based registration (NCC : 0.28) in red box region.}
\label{fig:pixel_comp}
\end{figure}

To compute similarities between adjacent EM sections, we generate data-driven features via a convolutional autoencoder, which consists of 1) a convolutional encoder comprised of convolutional layers with ReLU activations and 2) a deconvolutional decoder comprised of deconvolutional layers with ReLU activations that were symmetrical to the encoder without fully connected layers. 
%
%
Therefore, our method is applicable to any sized dataset (i.e., the network size is not constrained to the input data size). 
Our autoencoder can be formally defined as follows:
\begin{equation}
h = f_\theta(x)
 \label{eq:encoder}
\end{equation}
\vspace{-0.2in}
\begin{equation}
 y = g_\phi(h)
 \label{eq:decoder}
\end{equation}
\vspace{-0.2in}
\begin{equation}
 L_{\theta, \phi} = \sum_{i=1}^N ||x_i-y_i||^2_2 + \lambda(\sum_k ||\theta_k||^2_2 + \sum_k ||\phi_k||^2_2)
 \label{eq:autoencoder_loss}
\end{equation}
where $f_\theta$ and $g_\phi$ are the encoder and the decoder and $\theta$ and $\phi$ are their parameters, respectively.
The loss function (Eq.~\ref{eq:autoencoder_loss}) consists of the reconstruction term minimizing the difference between the input and output images and the regularization terms minimizing the $\ell_2$-norm of the weights of the network to avoid overfitting. 
Fig.~\ref{fig:pixel_comp} shows that our autoencoder feature-based registration generates more accurate results compared to the conventional pixel intensity-based registration, i.e., (d) shows the smaller normalized cross correlation (NCC) error between aligned images than (c). 
%
%

\vspace{-0.3in}
\subsection{Deformable Image Registration using a Spatial Transformer Network}
\begin{figure}[t]
	\centering
	\includegraphics[width=\linewidth]{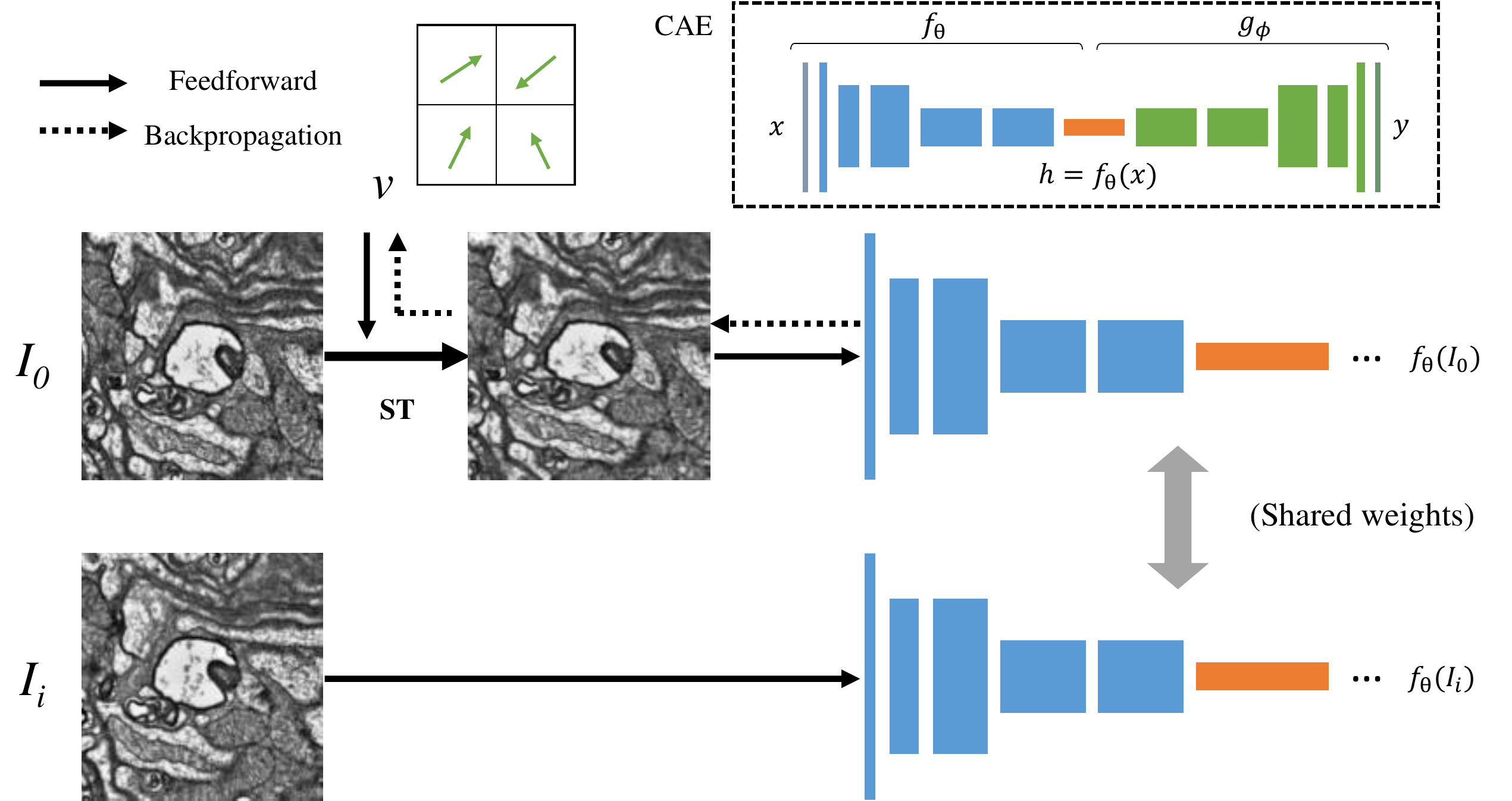}
	\caption{The overview of our method. The upper right dashed box represents the pretrained convolution autoencoder (CAE). The alignment is processed by backpropagation with loss of autoencoder features.}
\label{fig:overview}
\end{figure}

Upon completion of autoencoder training, a spatial transformer (ST) module $T$ is attached to the front half (i.e., encoder) to form the proposed spatial transformer network  (see Fig.~\ref{fig:overview}, refer to~\cite{jaderberg2015spatial} for the details of the ST module).
%
This design is intended to find the proper deformation of the input image via an ST by minimizing the registration error measured by the pre-trained autoencoder.
 The objective function for registration errors between the reference and the moving images is formulated as Eq.~\ref{eq:alignment_loss}. 
%
 The reference image $I_1$ is fixed, and the moving image $I_0$ is deformed by the ST with the corresponding vector map $v$. 
 %
Notably, the resolution of vector map $v$ is usually coarser than that of the input image. 
Therefore, we need smooth interpolation of a coarse vector map to obtain a per-pixel moving vector for actual deformation of the moving image. 
A thin plate spline (TPS) was used in the original STN for a smooth deformable transform, but other interpolation schemes, such as bilinear, bicubic, B-spline, etc., can be used as well. 
In our experiment, bilinear interpolation produced better results with finer deformation compared to the TPS.
 \begin{equation}
 L_{v}(I_0, I_1) = ||f_\theta(I_1) - f_\theta(T_v(I_0))||_2^2 + \alpha ||v||_2^2 + \beta ||\nabla v_x||_2^2 + \gamma ||\nabla v_y||_2^2
 \label{eq:alignment_loss}
\end{equation}

 The first term of Eq.~\ref{eq:alignment_loss} measures how two images are contextually different via a trained autoencoder. 
 We assumed that if the encoded features of two images are similar, then the images themselves are also similar and well-aligned. 
 %
 %
 %
 %
 %
The rest of the terms in Eq.~\ref{eq:alignment_loss} reflect the regularization of vector map $v$, which penalizes large deformation while promoting smooth variation of the vector map, and $\alpha$, $\beta$, $\gamma$ are their corresponding weights.
%
%
Because every layer is differentiable, including an ST, we directly optimize $v$ by backpropagation with a chain rule, in which only $v$ is updated and the weights in the autoencoder are fixed. 
We used the ADAM optimizer~\cite{kingma2014adam} for all our experiments.

The objective function using only adjacent image pairs could be vulnerable to imaging artifacts, which may result in drifting due to error accumulation when many sections are aligned.
%
To increase the robustness of alignment, we extend the objective function (Eq.~\ref{eq:alignment_loss}) to leverage multiple neighbor sections. 
%
%
Let the moving image be $I_0$, its neighbor reference $n$ images be $I_1$ to $I_n$, and their corresponding weights be $w_i$.
The proposed objective function (Eq.~\ref{eq:batch_loss}) combines the registration errors across neighbor images, which can lessen strong registration errors from images with artifacts and avoid large deformation.

To accumulate the registration error only within the image after deformation, we applied the \emph{empty space mask} that represents the empty area outside the image.
%
%
After image deformation, we collect the pixels outside the valid image region and make a binary mask image. 
%
We resize this mask image to match the size of the autoencoder feature map using a bilinear interpolation (shown as $M(T)$ in Eq.~\ref{eq:batch_loss}).
Based on this objective function, the alignment of many EM sections is possible in an out-of-core fashion using a sliding-window method.

\vspace{-0.2in}
\begin{equation}
 L_{v}(I_0, ..., I_n) = \sum_{i=1}^n w_i M(T_v) ||f_\theta(I_i) - f_\theta(T_v(I_0))||_2^2 + \alpha ||v||_2^2 + \beta ||\nabla v_x||_2^2 + \gamma ||\nabla v_y||_2^2
 \label{eq:batch_loss}
\end{equation}


We also developed a technique for handling images with dusts and folds. 
Because the feature errors are high in the corrupted regions, we selectively ignore such regions during the optimization, which we call~\emph{loss drop}.
%
%
%
This is similar to applying an empty space mask except that pixel selection is based on feature error. 
In our implementation, we first dropped the top $50\%$ of high error features, and then reduced the dropping rate by half per every iteration.
By doing this, we effectively prevented local minimums and obtained smoother registration results.

\vspace{-0.07in}

\section{Results}
\label{sec:result}

%

We implemented our method using TensorFlow, and used a GPU workstation equipped with an NVIDIA Titan X GPU.
 %
We used three EM datasets: transmission EM (TEM) images of \textit{Drosophila} brain, human-labeled TEM images of another \textit{Drosophila} brain provided by CREMI challenge\footnote{https://cremi.org/} (those two Drosophila images are collected independently on separate imaging systems), and mouse brain scanning EM (SEM) images with fold artifacts.
We used two convolutional autoencoders: one is a deeper network (as shown in Fig.~\ref{fig:overview}) with $3 \times 3$ filters used for the \textit{Drosophila} TEM datasets, and the other is a shallower network with a larger filter size (i.e., 6 layers with $7 \times 7$ filters) used for the mouse SEM dataset. 
In bUnwarpJ and elastic alignment experiments, we performed various experiments to find the optimal parameters and selected the parameters that gave the best results.

\begin{figure}[ht]
	\centering
	\includegraphics[width=\linewidth]{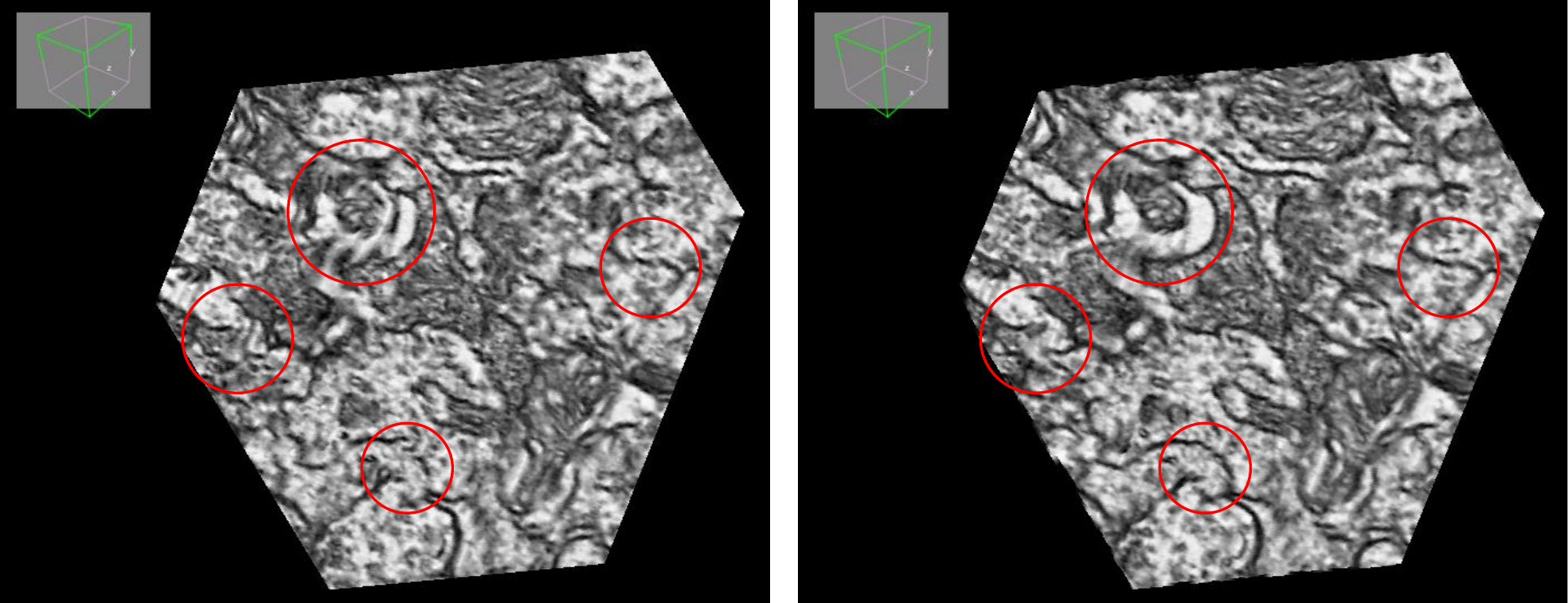}
	\caption{Drosophila melanogaster TEM dataset. Left : Pre-aligned result. Right : After registration using our method.}
	\label{fig:3D_cut}
\end{figure}

\vspace{-0.1in}
\paragraph{\textbf{Drosophila TEM data}}
The original volumetric dataset comprises the anterior portion of an adult female \textit{Drosophila melanogaster} brain cut in the frontal plane.
Each section was acquired at $4\times 4\times40$ nm$^3$vx$^{-1}$, amounting to 4 million camera images and 50 TB of raw data.
The original large-scale dataset was aligned with AlignTK (http://mmbios.org/aligntk-home) requiring extensive human effort and supercomputing resources. %
Although the alignment was sufficient for manual tracing of neurons, it must be improved for accurate and efficient automated segmentation approaches. 
Small volumes ($512\times512\times47$) were exported for re-alignment centered around synapses of identified connections between olfactory receptor neurons and second order projection neurons in the antennal lobe. 
 %
 %
 %
 %
 %
 Fig.~\ref{fig:3D_cut} shows the result of our registration method. 
Fig.~\ref{fig:3D_cut} left is the oblique (i.e., not axis-aligned) cross-sectional view of the original stack.
Due to inaccurate pre-alignment, some discontinuous membranes are shown (see the red circle areas), which are corrected in the aligned result using our method (Fig.~\ref{fig:3D_cut} right).

\begin{figure}[htb]
	\centering
	\includegraphics[width=1.0\linewidth]{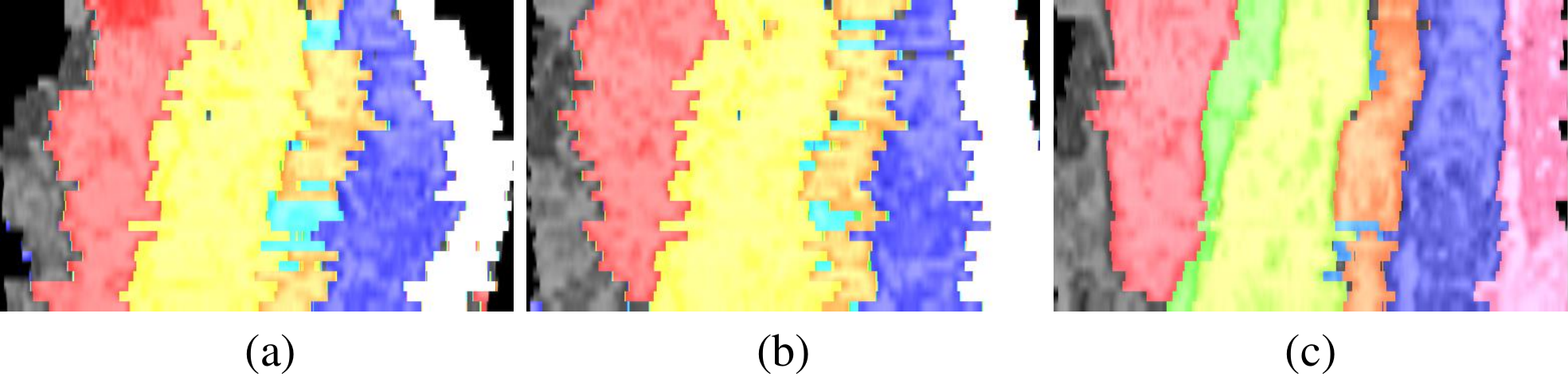}
	\caption{Vertical view of the alignment result of the randomly deformed CREMI dataset. (a) bUnwarpJ, (b) elastic alignment, and (c) our method. Each neuron is assigned a unique color.}
\label{fig:CREMI}
\end{figure}

\paragraph{\textbf{Labeled Drosophila TEM data from CREMI challenge}}
To quantitatively assess the registration quality, we used a small sub-volume ($512\times512\times31$) of registered and labeled TEM data from the CREMI challenge as a ground-truth.
We first randomly deformed both the raw and the labeled images using a TPS defined by random vectors on random positions. 
The random positions were uniformly distributed in space, and the random vectors were sampled from the normal distribution with a zero mean value. 
Then we performed image registration using three methods (bUnwarpJ, elastic alignment and our method).
Fig.~\ref{fig:CREMI} shows the vertical cross section of each result.
The bUnwarpJ result shows large deformation (i.e., drifting) across stacks (see the black regions on both sides). 
Although elastic alignment and our method show less deformation but our method clearly shows more accurate vertical membrane alignment.
To quantitatively measure the registration accuracy, we selected the 50 largest neurons and calculated the average Dice coefficient for each result, which came to 0.60, 0.73, and 0.83 for bUnwarpJ, elastic, and our method, respectively. 
This result shows that our registration method is more robust and resilient to random deformation.

\begin{figure}[ht]
	\centering
	\includegraphics[width=1.0\linewidth]{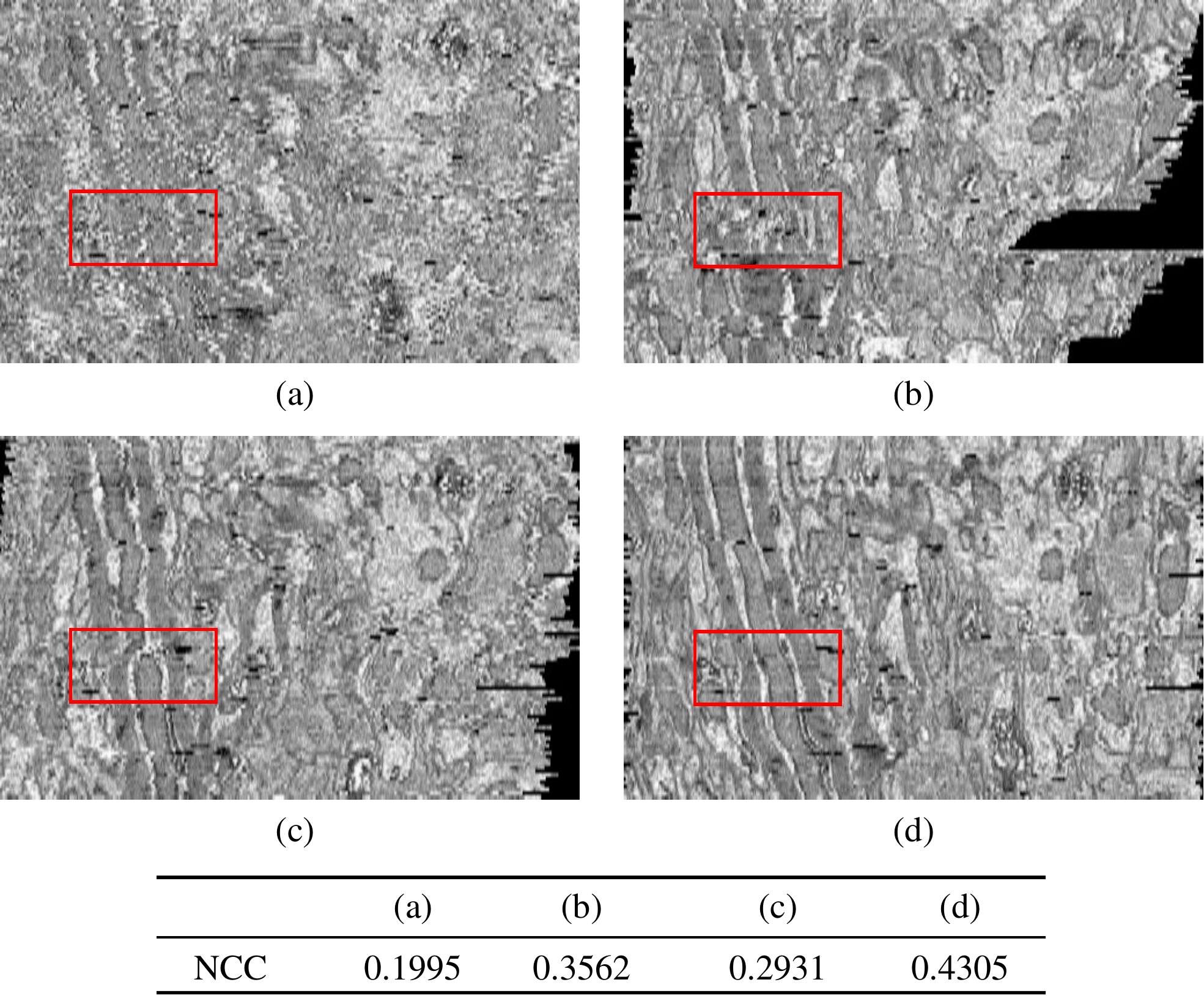}
	\caption{Visual comparison of mouse ssEM image registration results. (a) before alignment, (b) bUnwarpJ, (c) elastic alignment, and (d) our method. The red box is the region near the folds (shown as black spots). The below table shows NCC of the inner region in each aligned result (black backgroud regions are not counted for NCC computation)}. 
\label{fig:crack_reconsturction}
\end{figure}

\paragraph{\textbf{Mouse lateral geniculate nucleus SEM data with fold artifacts}}
We next sought to assess the applicability of our new alignment method to data acquired from different EM imaging methods, using different model organisms, and containing fold artifacts.
A small volume ($1520\times2500\times100$) was selected from a mouse lateral geniculate nucleus dataset generously provided by the Lichtman laboratory~\cite{morgan2016}.
This dataset was acquired using SEM with a resolution of $4\times4\times30$ nm$^3$vx$^{-1}$, and contains folds caused by cracks in the substrate onto which sections were collected.
Fig.~\ref{fig:crack_reconsturction} shows the vertical cross section of the registration results as compared to conventional registration methods. 
The overall registration quality of our method is higher than those of other methods, as indicated by clearer neuronal structures with low deformation. 
In particular, the red box shows a region containing warping due to folds, where our method is able to produce a smoother and more continuous result than others. 

\section{Discussion and Conclusion}
\label{conclusion}

One problem with the convolution operator is that it is neither scale- nor rotation-invariant.
We addressed this problem by generating features on the deformed image in every iteration and dynamically calculating feature differences.
Our method is slow due to the nature of learning algorithms, but the parameter tuning is much easier than existing methods, which makes it practically useful. 

In this paper, we introduced a novel deep network for ssEM image registration that is easier to use and robust to imaging artifacts. 
The proposed method is a general learning-based registration model that can easily extend to various applications by modifying the network. 
Improving running time via parallel systems and deploying our method on tera-scale EM stacks would be an interesting and important future research direction. 
We also plan to employ various interpolation schemes and feature encoding networks in the future. 

\vspace{0.05in}

\noindent
\textbf{Acknowledgements.}
This work is partially supported by the Basic Science Research Program through the National Research Foundation of Korea funded by the Ministry of Education (NRF-2017R1D1A1A09000841) and the Software Convergence Technology Development Program through the Ministry of Science, ICT and Future Planning (S0503-17-1007).

\vspace{-0.1in}

\bibliographystyle{splncs03}
\bibliography{paper60}

\end{document}